\pdfoutput=1
\documentclass[11pt]{article}

\usepackage{acl}
\usepackage{times}
\usepackage{latexsym}
\usepackage[T1]{fontenc}
\usepackage[utf8]{inputenc}
\usepackage{microtype}
\usepackage{amsmath}
\usepackage{graphicx}
\usepackage{booktabs}
\usepackage{enumitem}
\usepackage{titlesec}

\titlespacing\paragraph{0pt}{0.8pt}{1.6ex plus 0.1ex}

\title{Retrieval-Augmented Chain-of-Thought in Semi-structured Domains}

\author{Vaibhav Mavi \and Abulhair Saparov \and Chen Zhao \\
        vm2241@nyu.edu \\ as17582 \\ cz1285}

\author{Vaibhav Mavi \\
  New York University  \\
  \texttt{vm2241@nyu.edu} \\\And
  Abulhair Saparov \\
  New York University  \\
  \texttt{as17582@nyu.edu} \\\And
  Chen Zhao \\
  New York University  \\
  \texttt{cz1285@nyu.edu} \\}

\begin{document}
\maketitle
\begin{abstract}
Applying existing question answering (QA) systems to specialized domains like law and finance presents challenges that necessitate domain expertise. Although large language models (LLMs) have shown impressive language comprehension and in-context learning capabilities, their inability to handle very long inputs/contexts is well known. Tasks specific to these domains need significant background knowledge, leading to contexts that can often exceed the maximum length that existing LLMs can process. This study explores leveraging the semi-structured nature of legal and financial data to efficiently retrieve relevant context, enabling the use of LLMs for domain-specialized QA. The resulting system outperforms contemporary models and also provides useful explanations for the answers, encouraging the integration of LLMs into legal and financial NLP systems for future research.
\end{abstract}

\section{Introduction}
Building NLP systems for answering questions in the legal and financial domains could save time and resources, ensure compliance, and enhance the overall accuracy and effectiveness of legal and financial operations \cite{llmtax, fingpt}.
Applying QA systems to such domains poses unique challenges. These domains feature complex jargon, nuanced phrasing, and contextual dependencies that require specialized knowledge and expertise \cite{legalnlp1, bloomberggpt}. A system tailored to these domains should be able to efficiently process and analyze large volumes of legal, financial, or regulatory documents, extracting relevant insights and answering targeted queries.

\begin{figure*}
    \centering
    \vspace{-2em}
    \includegraphics[width=\linewidth]{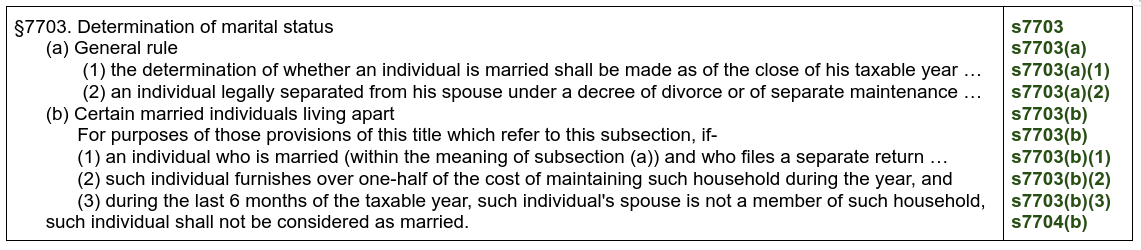}\vspace{-0.5em}
    \caption{An example of a statute from US Internal Revenue Code (left) and the subsection name assigned to each sentence after parsing as described in section~\ref{sec:sara_retrieval} (right).}
    \label{fig:sara_sec_eg}
\vspace{-1.2em}
\end{figure*}

Large language models (LLMs) have shown impressive performance on several NLP tasks \cite{surveyllm}. \citet{gptfinance} show that LLMs trained on large amounts of data are able to obtain the necessary domain knowledge through in-context learning (ICL) \cite{icl}. However, a major limitation of LLMs is the limit on the input size. There are many attempts to address this limitation \cite{alibi, nopos, pose} and multiple transformer models are able to handle longer contexts \cite{gpt4,codellama,xltransformer, long_context1}. However, \cite{lostmiddle} show that model performance on certain parts of the input decreases with input size. Further, the cost and latency of LLMs increases with the input size. 

The context required for legal and financial questions is often large and may not fit within the token limit, requiring more efficient retrieval. Financial and legal documents are often semi-structured. For example, Figure~\ref{fig:sara_sec_eg} shows a section from the US Internal Revenue Code. The text is organized into subsections, paragraphs and bullet points, which we leverage for better information retrieval. Further, financial reports often contain quantities in tabular format. We exploit these structures in a prompting approach that incorporates retrieval to workaround the context token limit.

We evaluate the proposed method on two datasets: FinQA \cite{finqa} and SARA \cite{sara}. These datasets feature complex questions which require multiple steps of reasoning and arithmetic computations, which is challenging for language systems.
We adopt chain-of-thought (CoT) prompting \cite{cot} for generating the answers since it is well suited for performing reasoning in a step-by-step manner. A chain of thought is a coherent sequence of reasoning steps that lead to the correct answer in a step-by-step manner. Providing examples of question-answer pairs along with their CoTs prepended to the test question causes GPT-3 to likewise output a CoT along with the answer for the test question, and improve its overall reasoning accuracy. CoT prompting is especially useful for complex tasks which require multiple steps of reasoning over the given input.

The results demonstrate that this simple and efficient approach outperforms state-of-the-art models in these domains. Training LLMs on financial and legal data may not be feasible as they may contain sensitive information. The use of ICL circumvents the problem and avoids expensive and tedious process of data collection and training. This makes the proposed approach a practical solution in scenarios where labeled data is limited or expensive to obtain. Additionally, CoT prompting offers the advantage of generating explanations and facilitating interpretability in critical domains where it is a key obstacle for the adoption of AI systems \cite{survey-explain}. However, a major drawback of the approach is that it is task-specific. In particular, the retrieval relies on the structure within in the data and needs to be adapted to data from different sources\footnote{The code for this work is publicly available at \href{https://github.com/vaibhavg152/Retrieval-Augmented-Chain-of-Thought-in-Semi-structured-Domains}{\texttt{github.com/vaibhavg152/Retrieval-Augmented-} \texttt{Chain-of-Thought-in-Semi-structured-Domains}}}.

We hope our work fosters research on coupling LLMs with retrieval in domains such as finance and law, where the ability to extract insights and answer questions about vast amounts of domain-specific data has many practical applications.

\section{Related work}
Previous work has proposed training specialized LLMs for financial and legal domains \cite{bloomberggpt, lawyerllama, lawgpt, fingpt}. However, doing so requires a large amount of data, compute and cost.

\citet{apollo} evaluate GPT-2 on FinQA \cite{finqa}. \citet{gptsara} evaluate GPT-3 with different prompting techniques on SARA where the context includes all the sections from the statutes. Since the input size of GPT-3 is limited, the prompts only included a subset of sections, which may not contain the required information. Further, fewer in-context examples were used for CoT as compared to few-shot learning. \citet{icl_num1} and \citealt{icl_num2} observe better performance with more in-context examples.

\citet{llmtax} test various GPT models with ICL to answer multiple choice questions over tax laws. A retriever augmented setting is tested where a dense passage retriver, GTR \cite{gtr}, retrieves the top 4 relevant sections to the questions. Since entire sections are passed to the LLMs, the text has to be truncated.

This study extends past work by complementing LLMs with a retriever that extracts the relevant text from within the statutes, allowing for larger contexts and more in-context examples in the prompt.

\begin{figure*}
    \centering
    \vspace{-2em}
    \includegraphics[width=\linewidth, height=0.83\columnwidth]{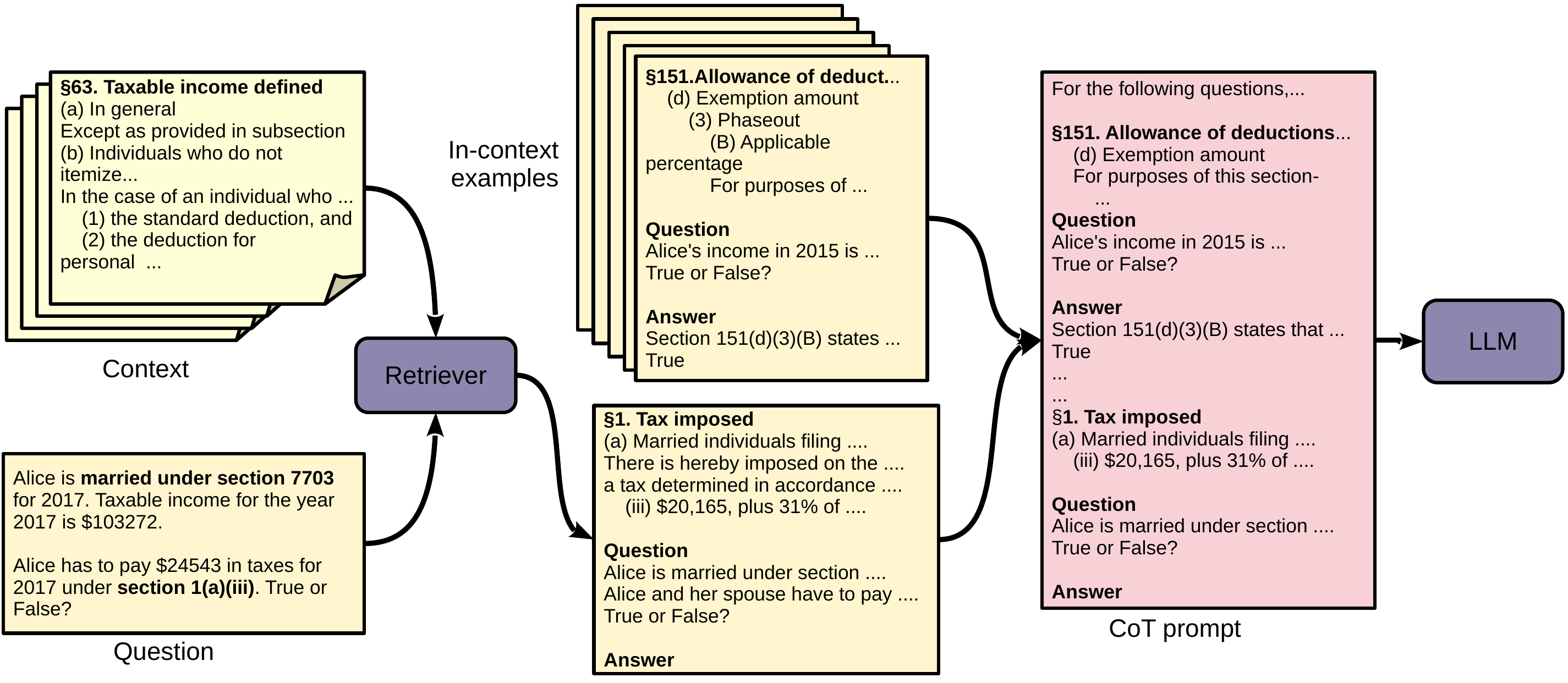}\vspace{-0.4em}
    \caption{An overview of the proposed system on a sample input from SARA. The retriever extracts the relevant information from the context and combines it with the question. In-context examples are appended with the retrieval output to construct a prompt which is used for querying LLMs to generate an answer along the chain-of-thought.}
    \label{fig:model_fig}
\vspace{-1.0em}
\end{figure*}

\section{Data}\label{sec:data}
We use two datasets containing questions that involve multi-step logical and arithmetic reasoning from the legal and financial domains respectively.
\subsection{SARA}
StAtutory Reasoning Assessment dataset (SARA) \cite{sara} is designed to evaluate statutory reasoning over a set of sections extracted from the US Internal Revenue Code (IRC). For each of the subsections contained in the selected sections, there are two hand-written case scenarios. 
Correctly solving these cases requires multiple steps of arithmetic as well as logical reasoning. For instance, some cases require computing the amount of tax owed according to a given section, only if the section applies to the given case. Thus, this dataset serves as a challenging task for an AI system, requiring domain expertise and reasoning abilities.

\subsection{FinQA}

FinQA \cite{finqa} is a financial QA dataset. 
It comprises of 8,281 examples where each question is accompanied by a financial report, containing text as well as a table. The report contains the necessary information to correctly answer the question. 
FinQA poses many challenges for a QA system. The questions require retrieval, arithmetic and logical reasoning simultaneously over tables and text. 
The questions also require an understanding of financial jargon. Finally, multiple reasoning steps are required to derive the answer.

\section{Methodology}\label{sec:method}
Figure~\ref{fig:model_fig} shows an overview of our proposed approach, which consists of two main components: retrieval and answering. The retrieval step involves filtering paragraphs from text and rows from tables that are relevant to the question. 
The retrieved information is then passed to the answering model.

\subsection{Retrieval}
Retrieval is essential for fully leveraging ICL and CoT reasoning abilities of LLMs. It can help to prevent the required context from exceeding the token limit, while also allowing the prompt to include enough in-context examples along with their CoT explanations. It can also help to reduce the time and cost of inference. 

We propose to leverage the structure present in the data to retrieve the relevant context from the legal statutes and financial reports. This structure is specific to the data source and the retriever needs to be designed accordingly. In our analysis, we explore datasets with two different sources: SARA, where a template-based algorithm can be used for effective retrieval; and FinQA, where a more sophisticated pre-trained retrieval model is required.

\subsubsection{SARA} \label{sec:sara_retrieval}
As shown in Figure~\ref{fig:sara_sec_eg}, the statutes in SARA are organized in a hierarchical structure with sections, sub-sections, paragraphs, and bullets. This hierarchical structure offers valuable information for efficient and accurate retrieval.

Figure~\ref{fig:sara_qn} in the Appendix shows an example of a question from SARA. The questions contain references to the specific sub-sections they pertain to. Firstly, a simple regular expression-based extractor scans the question text to identify the relevant section name.

Next, a rule-based statute parser extracts the mentioned sub-section. The parser reads each sentence in the given statutes and assigns it to the most specific sub-section to which the sentence belongs.
Figure~\ref{fig:sara_sec_eg} shows an example of a parsed statute section.
We explore three retrieval strategies: 
\begin{enumerate}[topsep=1.5pt,noitemsep,leftmargin=1.7em]
    \item \texttt{\textbf{mentioned-only}:} The retriever returns all the sentences that are assigned to sub-sections containing the queried sub-section as a prefix. For Figure~\ref{fig:sara_sec_eg}, a query for sub-section \textit{$7703(a)(1)$} will result in sentences assigned to $s7703, s7703(a)$ and $s7703(a)(1)$.
    \item \texttt{\textbf{entire-section}:} Retriever returns the entire sub-section. In Figure~\ref{fig:sara_sec_eg}, a query for sub-section \textit{$7703(a)(1)$} will result in sentences assigned to $s7703, s7703(a), s7703(a)(1),$ as well as $s7703(a)(2)$.
    \item \texttt{\textbf{references}:} Retriever returns sub-sections mentioned in the question along with those that are referenced in these retrieved subsections\footnote{It is intuitive to consider an approach that recursively retrieves text from sections mentioned in the sections retrieved in the previous step. However, this recursive approach proves to be impractical as it generates excessively large contexts.}.
\end{enumerate}

\subsubsection{FinQA}
The absence of a hierarchical structure in FinQA reports makes it impractical to adopt a rule-based approach for retrieval. \citet{finqa} convert the tables into text and then use BERT for retrieving relevant sentences from the report.

However, using templates to convert tables into text leads to very long contexts. These templates can also introduce grammatical and logical errors, leading to a loss in the performance of the answering module. Thus, we use a tabular format during the answering step in order to exploit the structure (see Figure~\ref{fig:finqa_prompt} in Appendix).

We also evaluate the system with gold retrieved sentences (GPT3-Gold, LLaMA2-70B-Gold).

\subsection{Answering}\label{sec:ans}

In this study, we test GPT-3 (\texttt{text-davinci-003}) \cite{icl} and LLaMA-2 \cite{llama2} to answer the queries. We experiment with different prompting techniques, namely \emph{zero-shot}, \emph{few-shot} and \emph{chain-of-thought} prompting. CoT prompting has been shown to improve the ICL abilities of sufficiently large LLMs \cite{surveyllm} and is especially useful for tasks that require multiple steps of reasoning.

In the zero-shot setting, the model is given the retrieved context and the question and is expected to output just the answer without any explanation.

In the few-shot setting, we further include in-context examples of question-answer pairs (8 examples for SARA and 12 examples for FinQA\footnote{Tabular data in FinQA leads to shorter retrieved context and allows more examples per prompt.}).

In the CoT setting, we use the same in-context examples as used for the few-shot setting but each example also includes a CoT explanation. These explanations are manually written for each example. The model is expected to generate the answer along with the CoT explanations for the test cases.

For all questions in a dataset, we use the same prompt containing the same in-context examples which are selected using prompt tuning as described in Appendix section~\ref{sec:tuning}.

Figures \ref{fig:sara_prompt} and \ref{fig:finqa_prompt} in the Appendix show the CoT prompts used for SARA and FinQA respectively.

\section{Experimental setup}\label{sec:eval}
\begin{table}
\centering
\footnotesize
{
\setcitestyle{square}
\setcitestyle{numbers}
\begin{tabular}{l | r }
\toprule
\textbf{Model name}     & \textbf{Accuracy} \\ \midrule
Majority baseline \cite{sara}       & $50.0 \pm 8.22$              \\
Feed-forward \cite{sara}            & $54.0 \pm 8.20$              \\
Legal-BERT \cite{sara}              & $49.0 \pm 8.22$              \\
BERT \cite{bertsara}                & $59.0 \pm 8.09$              \\
\midrule
GPT-3 (0-shot) \citep{gptsara}      & $71.0 \pm 7.46$              \\
GPT-3 (CoT) \cite{gptsara}          & $57.0 \pm 8.14$              \\
GPT-3 (dynamic) \cite{gptsara}      & $60.0 \pm 8.06$              \\
\midrule
GPT-3 + Ret                         & $\textbf{81.6}\pm \textbf{4.22}$   \\
LLaMA2-7B + Ret                     & $53.5 \pm 5.43$              \\
LLaMA2-7B\_chat + Ret               & $54.4 \pm 5.43$              \\
LLaMA2-13B + Ret                    & $57.5 \pm 5.39$              \\
LLaMA2-13B\_chat + Ret              & $66.7 \pm 5.13$              \\
LLaMA2-70B + Ret                    & $71.1 \pm 4.94$              \\ \bottomrule
\end{tabular}
}
\vspace{-0.5em}
\caption{Comparison of proposed system's performance on SARA with the existing baselines. The top section shows non-LLM based methods. The middle section shows the evaluation results from \citet{gptsara}. The bottom section shows the results of our proposed system with `Ret' representing the proposed retrieval. Results are shown with the $90\%$ confidence interval. }
\label{tab:sara_main}
\end{table}

\subsection{Evaluation}
For SARA, the task is formulated as an entailment task and is evaluated as a binary classification task.

For FinQA, 
\citet{finqa} propose \textbf{program accuracy} where the model is expected to generate a `program' along with the  answer. A program is a sequence of mathematical operations that leads to the final answer. 
The evaluation thus compares the output program with the gold standard program and checks if the two evaluate to the same answer.

We also measure the \textbf{answer accuracy} by ignoring errors only in units, prefix, suffix, precision digits or rounding errors.

\subsection{Comparison with existing methods}\label{sec:results}

\begin{table}
\centering
\footnotesize
\resizebox{\columnwidth}{!}
{%
\setcitestyle{square}
\setcitestyle{numbers}
\begin{tabular}{l|r|r}
\toprule
\textbf{Model}                      & \textbf{Program acc}                & \textbf{Answer acc}    \\ \midrule
Longformer \cite{longformer}        & 21.90 $\pm$ 2.01                    & 20.48 $\pm$ 1.96      \\
ELASTIC \cite{elastic}              & 57.54 $\pm$ 2.40                    & 62.16 $\pm$ 2.36      \\
DyRRen \cite{dyrren}                & 61.29 $\pm$ 2.37                    & 63.30 $\pm$ 2.34      \\
TabT5  \cite{tabt5}                 & \textbf{68.00} $\pm$ \textbf{2.27}  & \textbf{70.79} $\pm$ \textbf{2.21} \\
APOLLO \cite{apollo}                & 65.60 $\pm$ 2.31                    & 67.99 $\pm$ 2.27      \\
FinQANet-BERT \cite{finqa}          & 58.86 $\pm$ 2.39                    & 61.24 $\pm$ 2.37      \\ \midrule
GPT-3-BERT                          & \textbf{68.00} $\pm$ \textbf{5.43}  & 52.50 $\pm$ 5.81      \\
LLaMA2-7B-BERT                      & 25.50 $\pm$ 5.07	                  & 16.50 $\pm$ 4.32      \\
LLaMA2-7B\_chat-BERT                & 34.50 $\pm$ 5.53                    & 14.50 $\pm$ 4.10      \\
LLaMA2-13B-BERT                     & 52.50 $\pm$ 5.81                    & 33.00 $\pm$ 5.47      \\
LLaMA2-13B\_chat-BERT               & 50.50 $\pm$ 5.82                    & 26.50 $\pm$ 5.13      \\
LLaMA2-70B-BERT                     & 60.50 $\pm$ 5.69                    & 51.00 $\pm$ 5.81      \\ \midrule
Human non-expert \cite{finqa}       & 48.17 $\pm$ 2.43                    & 50.68 $\pm$ 2.43      \\
Human expert \cite{finqa}           & 87.49 $\pm$ 1.61                    & 91.16 $\pm$ 1.38      \\ \midrule
FinQANet-Gold \cite{finqa}          & 68.76 $\pm$ 2.25                    & \textbf{70.00} $\pm$ \textbf{2.23} \\
GPT-3-Gold                          & \textbf{72.50} $\pm$ \textbf{5.19}  & 56.50 $\pm$ 5.77      \\
LLaMA2-70B-Gold                     & 63.00 $\pm$ 5.62                    & 54.50 $\pm$ 5.79      \\ \bottomrule
\end{tabular}%
}
\vspace{-0.5em}
\caption{Comparison with state of the art and baselines methods on FinQA. Results are presented with a $90\%$ confidence interval.}
\label{tab:finqa_main}
\end{table}

Tables~\ref{tab:sara_main} and \ref{tab:finqa_main} show results on SARA and FinQA respectively\footnote{For testing on FinQA, we randomly sample 200 examples from the public test set due to the high cost of LLM queries.
}. Descriptions of the baselines are provided in Appendix Section~\ref{sec:baselines}.

On SARA, both GPT-3 and LLaMA2-70B surpass the existing methods by a significant margin. We also observe the expected trend of the performance improving with the increase in the model size, with GPT-3 (175B) performing significantly better with LLaMA-2 models \cite{llmscaling}\footnote{Although \citet{llama2} report similar performance of LLaMA-2 to GPT-3 on benchmark datasets, the task addressed here is domain-specific and requires more complex mathematical and logical reasoning than the benchmarks they use for evaluation.}.

On the other hand, the performance on FinQA with GPT-3 is comparable with baselines in terms of program accuracy but lags behind in answer accuracy. We believe this behavior is due to arithmetic errors made by LLMs \cite{llm-arithmetic}, resulting in cases with correct programs but incorrect answers.
Our approach with LLaMA2-13B/70B and GPT-3 outperforms general crowd workers who lack domain expertise in finance, whereas it falls short compared to financial experts.

The bottom section of Table~\ref{tab:finqa_main} highlights the effectiveness of GPT-3 over FinQANet \cite{finqa} when provided with the gold retrieved results. However, LLaMA-2 shows sub-par performance.

\subsection{Ablation studies}
\textbf{Comparison of prompting techniques:} Table~\ref{tab:prompt} in the Appendix shows the evaluation results for zero-shot, few-shot and CoT prompting. CoT prompting leads to significantly better results across all models.

\noindent \textbf{Comparing retrieval strategies:} As outlined in section \ref{sec:method}, we test three different retrieval strategies for SARA.
Table~\ref{tab:sara_abl} reveals that \texttt{mentioned-only} and \texttt{references} perform significantly better than \texttt{entire-section}. The questions in SARA are designed in a way where additional context apart from the mentioned sub-sections is not required. The difference in accuracy indicates the benefit of more targeted retrieval for model performance, since over-retrieval may dilute the signal provided by more directly relevant context.

\begin{table}
\centering
\footnotesize
{
\begin{tabular}{l r} \toprule
\textbf{Retrieval strategy} & \textbf{Accuracy}         \\ \midrule
\texttt{entire-section}              & 52.50 $\pm$ 12.99              \\
\texttt{references}                  & 75.00 $\pm$ 11.26              \\
\texttt{mentioned-only}              & \textbf{77.50} $\pm$ \textbf{10.86}         \\ \bottomrule
\end{tabular}
}
\vspace{-0.5em}
\caption{Comparison of the three retrieval strategies used with GPT-3 on SARA validation set.}
\label{tab:sara_abl}
\vspace{-1.4em}
\end{table}

\noindent \textbf{Case analysis}
We perform manual qualitative inspection of the generated CoT explanations and report the analysis in Appendix section \ref{sec:case}.

\section{Discussion}\label{sec:discuss}
This study aims to utilize LLMs for challenging domain-specific QA tasks by using ICL along with retrieval techniques that leverage the semi-structured nature of financial and legal data.
The proposed approach is simple and performs well compared to existing systems. It exploits ICL which avoids the costly and time-consuming processes of data collection and training. Since the proposed system produces a chain-of-thought with each output, it is easily interpretable and errors can be identified and rectified by human supervision \cite{survey-explain}.

We hope this work will encourage researchers to delve deeper into the analysis and development of LLM-integrated NLP systems and retrieval-augmented LLMs.

\section{Limitations}
The retrieval algorithms in our study are specifically tailored to each dataset. Despite good reasoning abilities, the evaluation reveals that arithmetic errors are common. Further, inference with LLMs can be costly with latency higher than traditional approaches, making it sub-optimal for handling large volumes of data efficiently.

These limitations point to interesting future directions such as using arithmetic tools as plugins \cite{toolformer} for better performance and more generalizable retrieval algorithms. Further, several domain-specific LLMs can be tested \cite{lawyerllama, fingpt, bloomberggpt}.


\section{Acknowledgements}
We sincerely thank the anonymous reviewers for their valuable feedback. For the experiments with LLaMA2, we thank the NYU IT High Performance Computing resources, services, and staff expertise.

\bibliography{anthology,custom}

\appendix

\section{Appendix}
\label{sec:appendix}

\subsection{Prompt tuning}\label{sec:tuning}
We iteratively refine the prompt using the validation sets of 40 samples for each dataset, with the aim of finding a prompt that encompasses a diverse range of cases while avoiding an overabundance of trivial or similar examples.

\section{Baselines} \label{sec:baselines}
On SARA, we evaluate our system against the following baselines:
\begin{itemize}
    \item \textbf{Majority baseline:} A trivial baseline that predicts the majority class for all the samples.
    \item \textbf{Feed-forward:} The feed-forward networks evaluated by \citet{sara}.
    \item \textbf{Legal-BERT:} A BERT model trained specifically on legal domain \cite{legalbert} and adapted for SARA by \citet{sara}.
    \item \textbf{BERT:} A BERT model adopted for SARA by \citet{bertsara}
    \item \textbf{GPT-3 (0-shot):} GPT-3 evaluated with a 0-shot prompt and without retrieval \cite{gptsara}.
    \item \textbf{GPT-3 (CoT):} GPT-3 evaluated with a CoT prompt and without retrieval \cite{gptsara}.
    \item \textbf{GPT-3 (dynamic):} GPT-3 evaluated with a dynamic few-shot prompt and without retrieval\cite{gptsara}. The prompt includes different in-context examples for different questions.
\end{itemize}
On FinQA, we compare our system with the following baselines in Table~\ref{tab:finqa_main}:
\begin{itemize}
    \item \textbf{Pre-trained Longformer:} Longformer (\citealp{longformer}) is a model designed to take long input documents in one step. The model can be seen as a representative of one-step approaches.
    \item \textbf{FinQANet:} \cite{finqa} use an LSTM decoder with attention for implementing the program generator and different models of BERT for retrieval.
    \item \textbf{ELASTIC:} \cite{elastic} use an adaptive symbolic compiler to generate the program.
    \item \textbf{DyRRen:} \cite{dyrren} employ dynamic reranking of retrieved facts in every step.
    \item \textbf{TabT5:} \cite{tabt5} use a T5 model pre-trained on the Wikipedia tables.
    \item \textbf{APOLLO:} \cite{apollo} The retriever is based on the sequence-pair classification following \cite{bert-passage}. The program generator leverages a BERT encoder and an LSTM decoder with attention mechanism along with consistency-based reinforcement learning.
\end{itemize}

\begin{figure}[!h]
    \centering
    \includegraphics[width=.9\columnwidth]{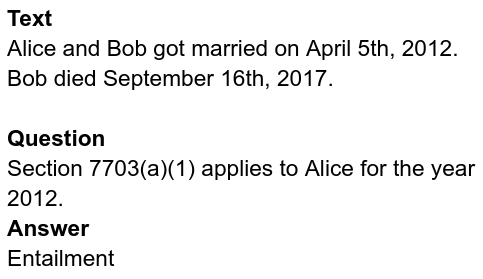}
    \caption{A case in SARA for section 7703.}
    \label{fig:sara_qn}
\end{figure}

\section{Data examples}
Figure \ref{fig:sara_qn} shows a question from SARA.

\section{Ablation studies}

\subsection{Zero-shot, few-shot and CoT prompting}
Table \ref{tab:prompt} shows the performance of different models with different prompting techniques. 
\begin{table}
\centering
\footnotesize
\resizebox{\columnwidth}{!}
{%
\begin{tabular}{l|c|c|c} \toprule
\textbf{Model}      & \textbf{Zero-shot} & \textbf{Few-shot}  & \textbf{CoT}        \\ \midrule
LLaMA2-7B           & 50.4 $\pm$ 5.4    & 58.8 $\pm$ 5.3    & 53.5 $\pm$ 5.4     \\
LLaMA2-7B\_chat     & 42.1 $\pm$ 5.3    & 51.8 $\pm$ 5.4    & 54.4 $\pm$ 5.4     \\
LLaMA2-13B          & 43.9 $\pm$ 5.4    & 53.1 $\pm$ 5.4    & 57.5 $\pm$ 5.4     \\
LLaMA2-13B\_chat    & 60.1 $\pm$ 5.3    & 53.9 $\pm$ 5.4    & 66.7 $\pm$ 5.1     \\
LLaMA2-70B          & 49.6 $\pm$ 5.4    & 67.5 $\pm$ 5.1    & 71.1 $\pm$ 4.9     \\
GPT-3               & 64.9 $\pm$ 5.2    & 74.6 $\pm$ 4.7    & 81.6 $\pm$ 4.2     \\ \bottomrule
\end{tabular}%
}
\vspace{-0.5em}
\caption{Comparison of different prompting techniques used on SARA. The prompts used are as described in section \ref{sec:ans} and shown in Appendix section \ref{sec:prompts}. Note that the results shown here are from our proposed retrieval-augmented method, and not the same as the baselines shown in Table~\ref{tab:sara_main}, which come from \citet{gptsara}.}
\label{tab:prompt}
\end{table}

\subsection{Case analysis}\label{sec:case}
\paragraph{SARA} We conducted a manual analysis of the model's output on the validation set. Table~\ref{tab:sara_case1} presents the results of this analysis on SARA, indicating the number of examples where both the answer and the chain-of-thought reasoning provided by the model were correct, both were incorrect and cases where one of them was incorrect. We found that in 58.5\% of the examples, the model accurately predicted both the output and the reasoning. For the remaining cases, we categorized the errors into four distinct categories, shown in Table~\ref{tab:sara_case2}.
\begin{table}
\centering

\begin{tabular}{l | c c} \toprule
\textbf{}                       & \textbf{Correct CoT}  & \textbf{Incorrect CoT} \\ \midrule
\textbf{Correct ans}         & 23                    &  8                     \\
\textbf{Incorrect ans}       & 3                     &  6                     \\ \bottomrule             
\end{tabular}
\caption{Results of the manual analysis performed on the validation set using GPT-3.}
\label{tab:sara_case1}
\end{table}

\begin{table}
\centering
\begin{tabular}{l|c} \toprule
\textbf{Reasoning Error type} & \multicolumn{1}{l}{\textbf{\# of cases}} \\ \midrule
Arithmetic errors             & 8                                    \\
Logical errors                & 6                                    \\
Context too long              & 2                                    \\
Retrieval error               & 1                                    \\ \bottomrule
\end{tabular}
\caption{Error analysis on the validation set of SARA.}
\label{tab:sara_case2}
\end{table}

\paragraph{FinQA}
On the constructed validation set comprising 40 samples, we observe that 30 samples have correct answers as well as corresponding programs. For the remaining 10 samples, we manually classify the errors into different categories, as shown in Table~\ref{tab:finqa-case}.
\begin{table}
\centering
{%
\begin{tabular}{l | r} \toprule
\textbf{Error type} & \multicolumn{1}{l}{\textbf{\# of cases}} \\ \midrule
Arithmetic error    & 3                                    \\
Logical error       & 2                                    \\
Annotation error    & 2                                    \\
Retrieval error     & 3                                    \\ \bottomrule
\end{tabular}%
}
\caption{Error analysis of the incorrect examples on 40 samples from FinQA.}
\label{tab:finqa-case}
\end{table}

\section{Prompts}\label{sec:prompts}
\begin{figure*}
    \centering
    \includegraphics[width=\textwidth]{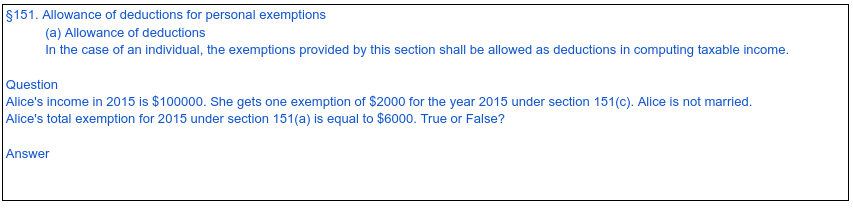}
    \caption{Chain-of-thought prompt for SARA for a sample. The complete prompt contains 8 in-context examples with CoT explanations followed by the question that the model is supposed to answer. The in-context examples and explanations remain the same for all questions in the dataset. The text highlighted in yellow are the CoT explanations that we hand-crafted, while the test question is shown in blue.}
    \label{fig:sara_prompt}
\end{figure*}

\begin{figure*}
    \centering
    \includegraphics[width=.97\textwidth]{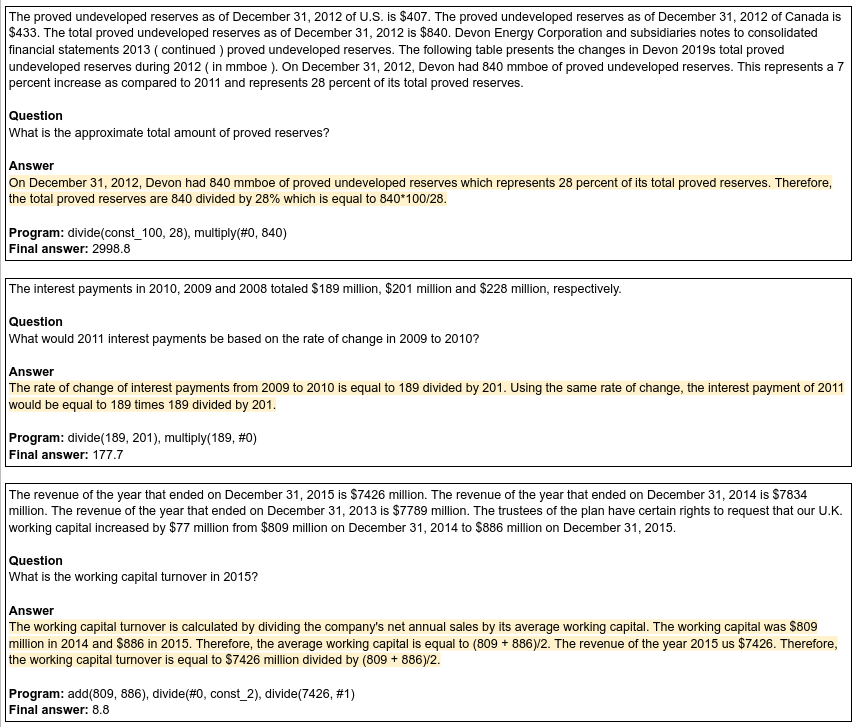}
    \includegraphics[width=.97\textwidth]{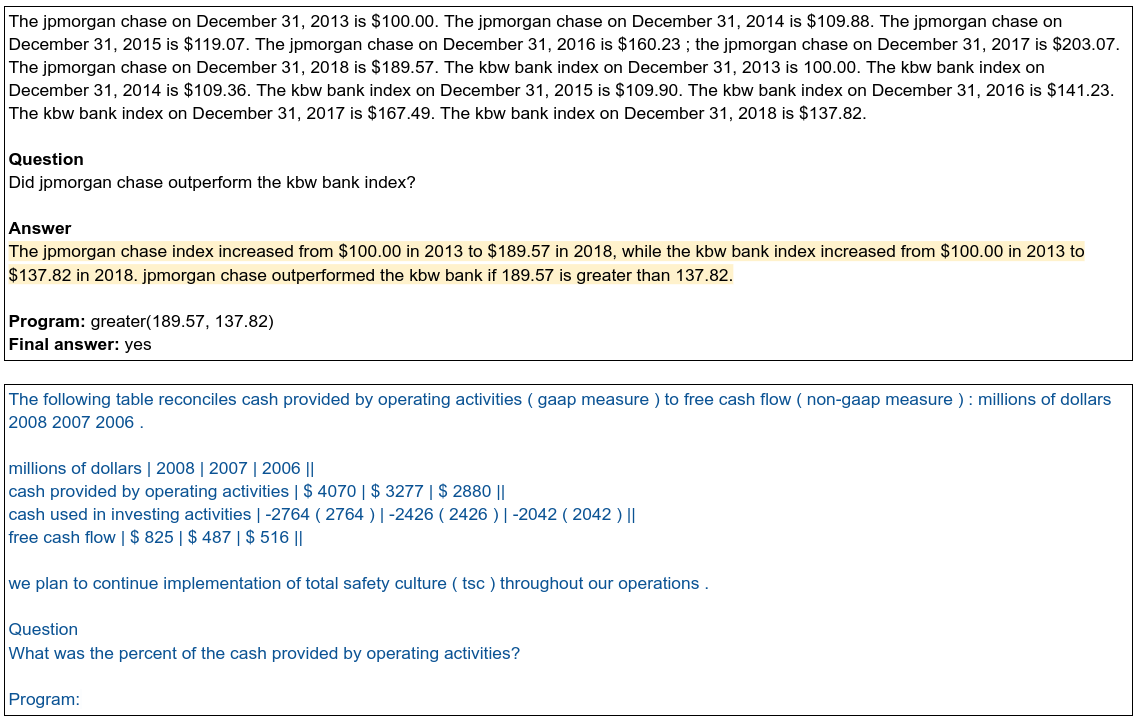}
    \caption{Chain-of-thought prompt for FinQA for a sample. The complete prompt contains 12 in-context examples with CoT explanations followed by the question that the model is supposed to answer. The in-context examples and explanations remain the same for all questions in the dataset. The text highlighted in yellow are the CoT explanations that we hand-crafted, while the test question is shown in blue.}
    \label{fig:finqa_prompt}
\end{figure*}

Figures \ref{fig:sara_prompt} and \ref{fig:finqa_prompt} show the prompts used for SARA and FinQA respectively.

\begin{figure*}
    \centering
    \includegraphics[width=\textwidth]{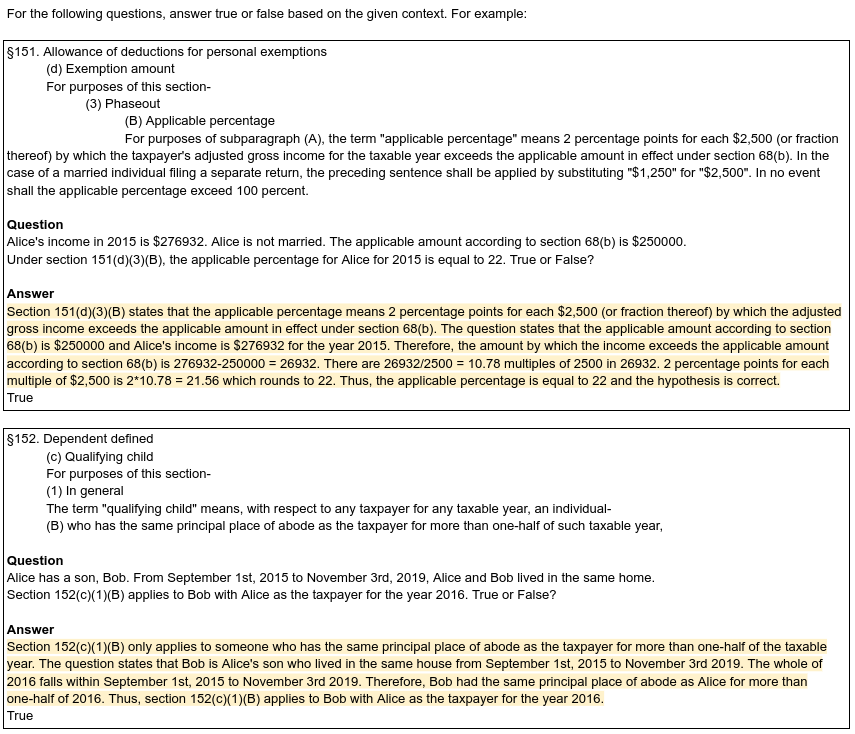}
    \includegraphics[width=\textwidth]{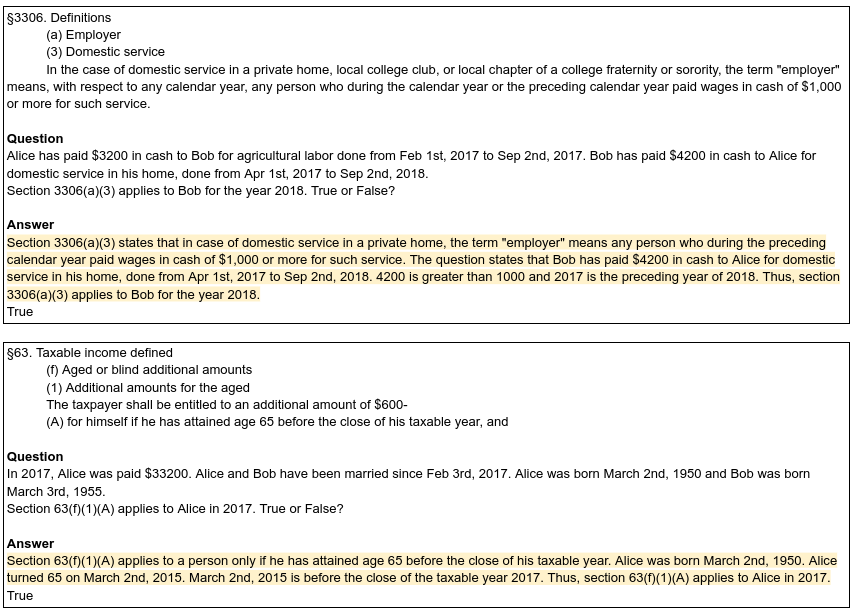}
\end{figure*}
\begin{figure*}
    \centering
    \includegraphics[width=\textwidth]{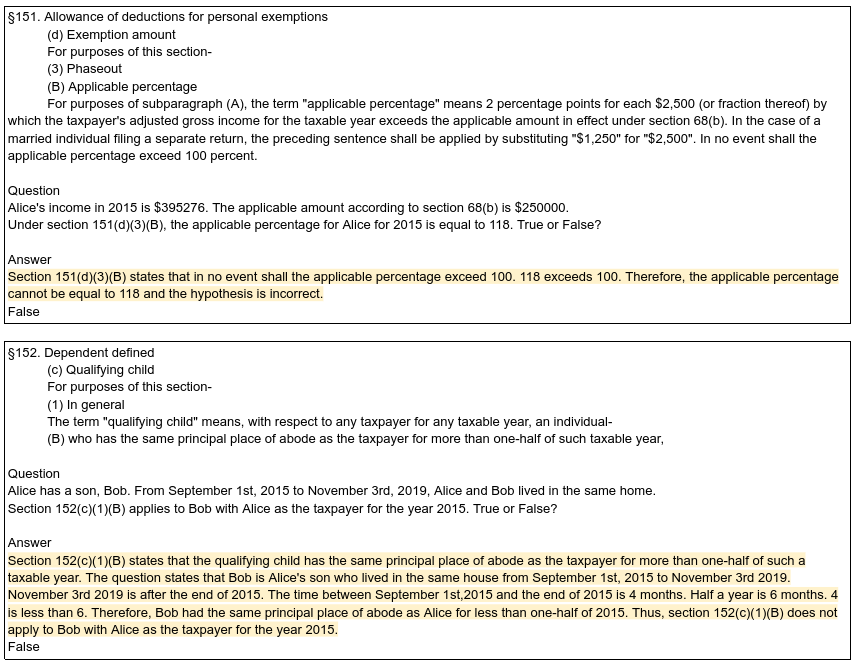}
    \includegraphics[width=\textwidth]{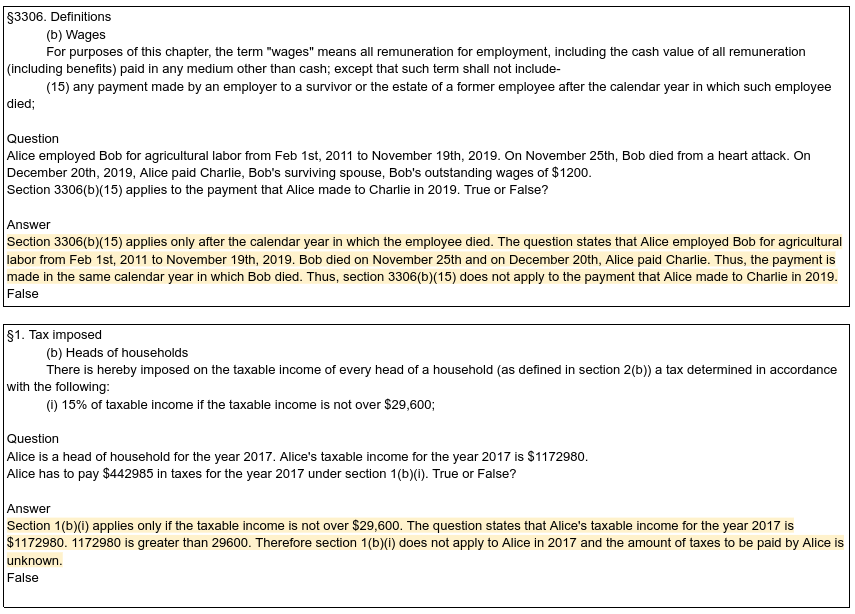}
\end{figure*}

\begin{figure*}
    \centering
    \includegraphics[width=\textwidth]{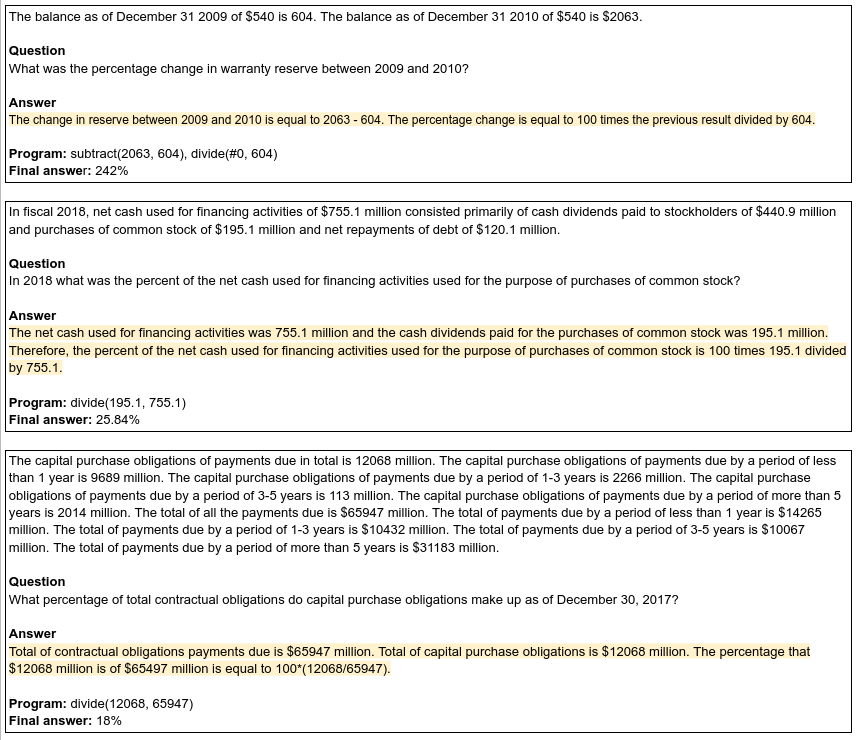}
\end{figure*}
\begin{figure*}
    \centering
    \includegraphics[width=\textwidth]{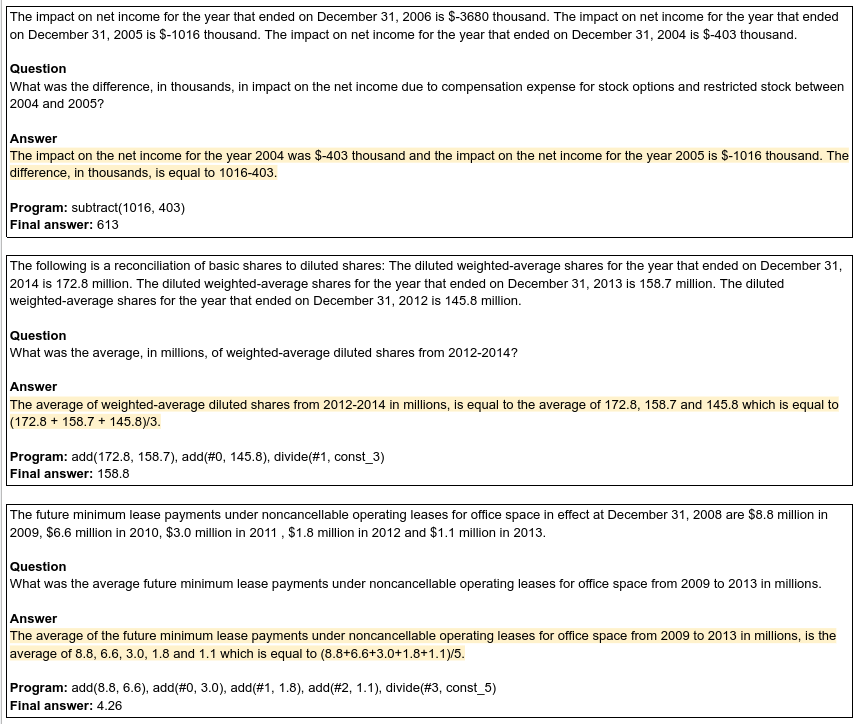}
    \includegraphics[width=\textwidth]{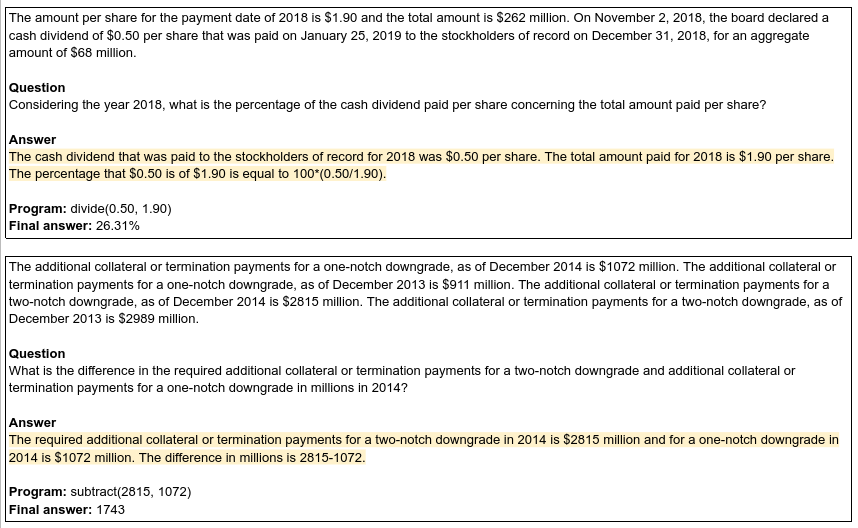}
\end{figure*}

\end{document}